\documentclass[10pt,twocolumn,letterpaper]{article}
\usepackage[accsupp]{axessibility} 
\usepackage{cvpr}              

\usepackage{multirow}
\usepackage{array}
\usepackage{booktabs}
\usepackage{colortbl}
\usepackage{xcolor}
\usepackage{amsmath}
\usepackage{enumitem}
\usepackage{subcaption} 
\definecolor{grassgreen}{rgb}{0.4, 0.8, 0.4}

\newcommand{\TODO}[1]{\textbf{\color{red}[TODO: #1]}}
\renewcommand{\TODO}[1]{}

\definecolor{cvprblue}{rgb}{0.21,0.49,0.74}
\usepackage[pagebackref,breaklinks,colorlinks,allcolors=cvprblue]{hyperref}

\def\paperID{14023} 
\def\confName{CVPR}
\def\confYear{2025}

\title{Unveiling Visual Perception in Language Models: \\An Attention Head Analysis Approach}

\author{
Jing Bi$^{1}$, Junjia Guo$^{1}$, Yunlong Tang$^{1}$, Lianggong Bruce Wen $^{2}$,
Zhang Liu$^{2}$, Chenliang Xu$^{1\dagger}$ \\ \\
$^{1}$University of Rochester, $^{2}$Corning Inc. \\ \\
{\tt\small \{jing.bi, yunlong.tang, chenliang.xu\}@rochester.edu} \\
{\tt\small \{jguo40\}@ur.rochester.edu} \\
{\tt\small \{Wenlb, liuz2\}@corning.com} \\
}
\begin{document}
\maketitle
\begin{abstract}
Recent advancements in Multimodal Large Language Models (MLLMs) have demonstrated remarkable progress in visual understanding. This impressive leap raises a compelling question: how can language models, initially trained solely on linguistic data, effectively interpret and process visual content? This paper aims to address this question with systematic investigation across 4 model families and 4 model scales, uncovering a unique class of attention heads that focus specifically on visual content. Our analysis reveals a strong correlation between the behavior of these attention heads, the distribution of attention weights, and their concentration on visual tokens within the input. These findings enhance our understanding of how LLMs adapt to multimodal tasks, demonstrating their potential to bridge the gap between textual and visual understanding. This work paves the way for the development of AI systems capable of engaging with diverse modalities.
\end{abstract}
\vspace{-6mm}
    
\section{Introduction}
\label{sec:intro}

The integration of large language models (LLMs) into multimodal tasks has led to impressive advancements, with MLLMs, or large vision-language models (LVLMs), achieving notable success in areas ranging from traditional image captioning to more complex visual dialogues.
A key element in training these models is the adapter, which connects the visual encoder and the LLM. The standard approach typically begins by freezing both the visual encoder (often based on models like CLIP~\citep{clip}) and the LLM, while fine-tuning only the adapter to map visual embeddings into a language-compatible space. 
Subsequently, the LLM is fine-tuned to follow instructions for visual tasks, enabling it to become a versatile model capable of solving vision problems that traditionally required multiple specialized models~\citep{bi2025verifybenchmarkvisualexplanation}.
This approach works remarkably well, raising an intriguing question: does the adapter translate visual embeddings into a language space that the LLM can interpret?

Recent studies~\citep{fromcliptodino, yang2024law} reveal that representations from encoders like CLIP can be effectively transformed into a space compatible with LLMs by training an adapter, whereas other visual representations face challenges in achieving similar results.
This raises a fundamental question about how LLMs, trained exclusively on language embeddings, interact with embeddings originating from fundamentally different visual spaces. 
In particular, \citet{chan2024analyzing} emphasizes that the embedding spaces for visual tokens differ significantly from those used for language, as visual tokens do not function as direct counterparts to language tokens but instead occupy a distinct space. 
Therefore, an in-depth analysis of how LLMs process embeddings from such a divergent space is essential, as current research lacks a clear understanding of how LLMs attend to these visual tokens during inference. 
To address this problem, it is necessary to explore the attention mechanisms of LLMs when handling visual tokens.
 
In parallel, research within the LLM domain has increasingly focused on uncovering the internal mechanisms that drive LLM capabilities, particularly around specialized attention heads that contribute distinct functions. 
For example, “Duplicate Heads” enhance the model’s ability to focus on repeated tokens, reinforcing patterns in textual data~\citep{wang2022interpretability}. Similarly, “Rare Words Heads” prioritize unique or low-frequency tokens~\citep{voita2019analyzing}, while “Previous Heads” capture positional relationships between sequential tokens~\citep{olsson2022context, ferrando2024information}. 
Another key attention mechanism, “Retrieval Heads,” boosts the model’s accuracy in locating specific tokens within long passages, a capability often called “Needle-in-a-Haystack”~\citep{wu2024retrieval}. 
Together, these examples demonstrate the complex internal structures within LLMs that support a wide range of linguistic functions. 
However, while much of the focus has been on these attention heads in the context of language, our work shifts the focus to understanding how LLMs process and attend to visual tokens.
\begin{figure*}[ht!]
  \centering
  \includegraphics[width=\textwidth]{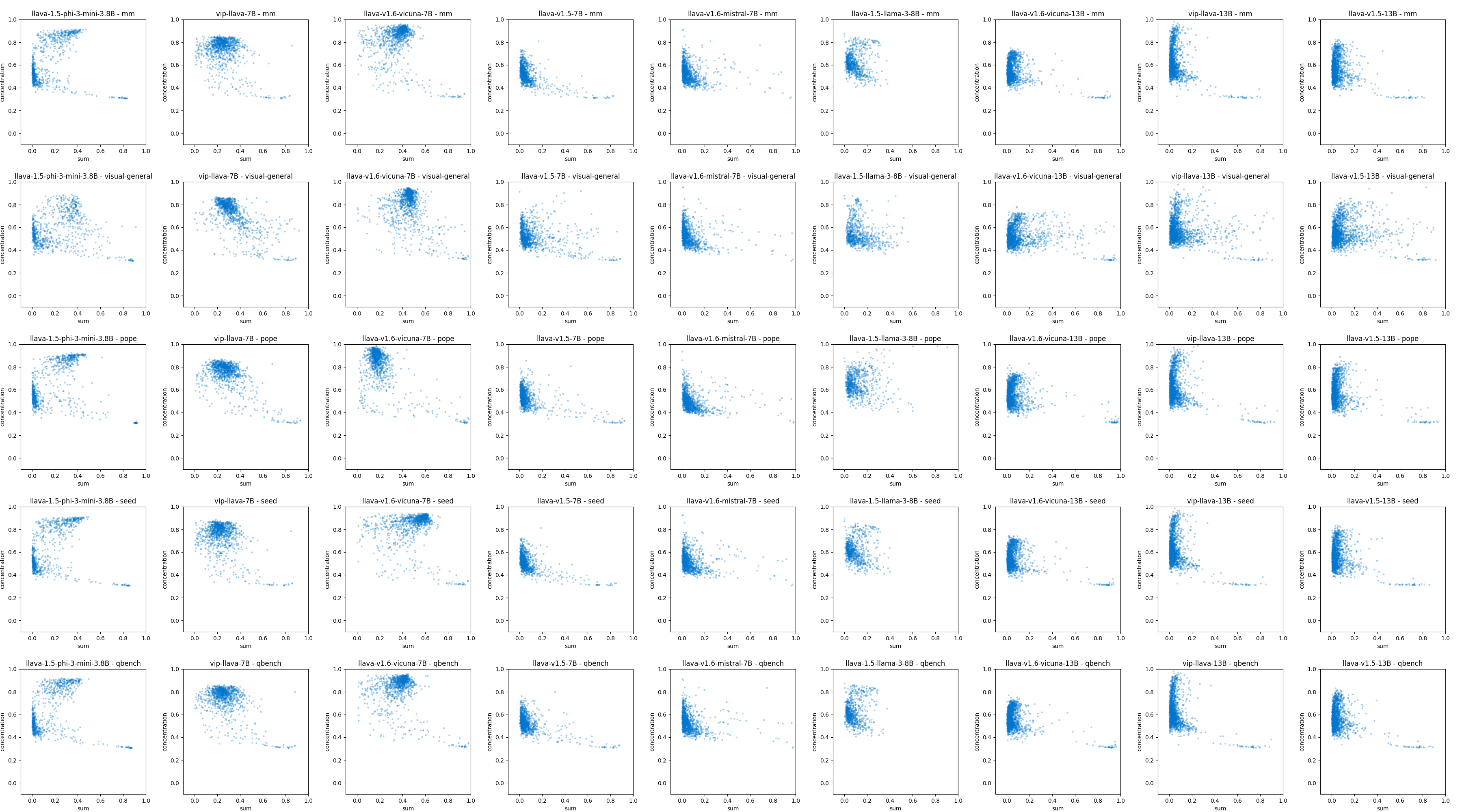}
  \caption{Attention vs Concentration across Different Models and Datasets:  Models are represented on the horizontal axis, and datasets are shown on the vertical axis within each subplot, with points representing individual attention heads. The x-axis represents the total attention weight assigned to visual tokens. The y-axis indicates the concentration of attention weights, where higher values signify focused attention on specific areas.The digits at the end of each subplot title represent the model’s accuracy under that dataset. Attention heads tend to fail under conditions of high content complexity, high weight, or both. Notably, attention distribution for each model remains consistent across different datasets, especially for the 13B model, which demonstrates low variation across datasets and models. This suggests that the proposed metrics—total attention weight and concentration—are reliable indicators of consistent model behavior across diverse datasets. }
  \label{fig:sumvcon}
    \vspace{-5mm}
\end{figure*}
In this work, we conduct extensive experiments across \textbf{4} model families, \textbf{4} model scales, and different training strategies.

Our investigation into visual heads is motivated by a key question: what exactly is the attention mechanism doing when processing visual tokens? To address this, we focused on a specialized dataset and designed a series of experiments to identify visual heads within popular LLM architectures. By leveraging a tailored dataset optimized for head detection, we conducted a comprehensive set of experiments across nine different models, rigorously analyzing the properties and behaviors of these visual heads in diverse contexts.
Our findings show that specific attention layers (varying by model) contain specialized “visual heads” that reliably focus on image tokens during task-solving, uncovering a unique mechanism through how LLMs process and interact with visual information. 
Unlike retrieval heads, which are universally distributed across LLM layers, visual heads tend to cluster within particular layers, indicating a unique structural adaptation for managing visual data.
This study not only sheds new light on the multimodal capabilities of LLMs but also paves the way for a deeper understanding of how their attention mechanisms operate across different representational spaces. Our findings reveal several distinct characteristics of visual heads:

\noindent\textbf{Gather in certain layers}: Unlike retrieval heads, which appear more universally across layers, visual heads tend to aggregate within specific layers, indicating a unique structural role. This was observed consistently across various model families (LLaMA 2~\citep{2023llama2}, Phi~\citep{abdin2024phi}, LLaMA 3~\citep{2024llama3}, and Mistral~\citep{jiang2023mistral}), scales (3.8B, 7B, 8B, and 13B).

\noindent\textbf{Concentration}: Our initial attempt at analyzing attention weights performed well on a controlled dataset; however, we found that it was inconsistent across different benchmarks. After in-depth analysis with Logic Lens\cite{belrose2023eliciting}, we identified that the concentration, or the sharpness, of the attention heads plays a crucial role in providing a consistent view of the model’s behavior across datasets.

\noindent\textbf{Model-Specific Activation}: The location and behavior of visual heads vary across models and training strategies~\citep{naveed2023comprehensive}, showing that visual heads are not static but adapt dynamically according to training conditions and contexts.

\noindent\textbf{Contextual Responsiveness}: Visual heads exhibit dynamic activation based on context, whether visual or textual, enabling them to adjust their function and behavior in response to varying visual and linguistic inputs, improving task performance and adaptability~\citep{de2023visual}.

\noindent\textbf{Enhanced Inference Efficiency:} Visual heads offer the potential for efficient inference, as only a small subset of heads is actively engaged in processing image tokens, which are often dominant in visual tasks. This discovery could lead to optimized models that activate attention heads for visual information, reducing computational load and speeding up inference times. Additionally, our findings explain the effectiveness of certain visual-token compression methods. Specifically, after the layers of the LLM, attention becomes concentrated on just a few focal areas of the input visual tokens, even if the input consists of thousands of tokens, with the remaining tokens receiving little to no attention.
\section{Related Work}

\subsection{MLLM Architecture}
MLLM often consist of modality encoders, an LLM, and an adapter to bridge these components.
Modality encoders serve to compress data, such as images or audio, into more concise representations. A widely adopted approach is to leverage pretrained encoders that have been pre-aligned with other modalities. For example, the CLIP~\citep{clip} model incorporates a visual encoder that has undergone extensive pretraining with text, establishing a semantic alignment between visual and textual modalities. This pre-aligned structure facilitates seamless integration with LLMs.
The LLaMA series~\citep{2023llama, 2023llama2, 2024llama3} and the Vicuna family~\citep{chiang2023vicuna,Bi_2024} are prominent open-source LLMs that have garnered significant academic interest. Additionally, increasing the parameter size of LLMs can yield further performance gains, much like enhancing input resolution. \citet{liu2023llava, liu2023improvedllava, liu2024llavanext,tang2023video}, observe that scaling an LLM from 7B to 13B parameters leads to notable improvements across various benchmarks. Moreover, when scaled up to 34B parameters, the model exhibits emergent zero-shot capability in Chinese, despite being trained solely on English  data.
\subsection{Learnable Connector}
This module plays a critical role in bridging the gap between different modalities, facilitating the efficient projection of information into a space that LLMs can interpret. Multimodal information fusion can be broadly categorized into two implementation strategies: token-level and feature-level fusion.
In token-level fusion, features output by encoders are transformed into tokens and concatenated with text tokens before entering LLMs. A common and practical approach is to employ a set of learnable query tokens to extract relevant information in a query-driven manner~\citep{carion2020end}, first introduced in BLIP-2~\citep{li2023blip} and subsequently adopted by numerous works~\citep{x-llm, instructblip, video-llama}. This Q-Former-style method condenses visual tokens into a limited number of representation vectors. 
Conversely, some methods opt for a simpler MLP-based interface to address the modality gap~\citep{liu2023llava, pmc-vqa, pandagpt, detgpt}. 
For instance, the LLaVA series leverages one or two linear MLPs~\citep{liu2023llava, liu2023improvedllava} to project visual tokens and align the feature dimensions with a space compatible with LLMs.

\subsection{Attention Head in LLM}
Large language models (LLMs) leverage a network of specialized attention heads to enhance their reasoning and interpretative abilities. 
These attention heads are designed to focus on different aspects of input data, processing information in ways that approximate human cognitive tasks. 
For instance, the Previous Head is designed to attend to the relationship between current and prior tokens, capturing token sequence patterns that support in-context learning~\citep{olsson2022context}. The Rare Words Head highlights infrequent tokens, aiding models in identifying unique and meaningful details within text~\citep{voita2019analyzing}. Syntactic Heads identify structural components such as subjects and objects, organizing information by grammatical roles for a more coherent understanding~\citep{ferrando2024information}. Additionally, Truthfulness Heads focus on maintaining the factual accuracy of responses, helping the model to produce reliable answers in response to questions~\citep{liang2024internal}.

To uncover these attention heads, researchers employ several different methods. Modeling-Free methods, such as zero ablation~\citep{pochinkov2024investigating} and activation patching~\citep{zheng2024attention} involve modifying latent states and observing their impact on model outputs without training. In contrast, Modeling-Required methods train additional models to assess head functions. These paper primarily emphasizes score-based methods, a Modeling-Free approach that uses mathematical scores, such as Retrieval~\citep{wu2024retrieval} and Negative~\citep{yu2024correcting} Attention Scores, to evaluate specific attention head behaviors directly. This focus on score-based methods provides an efficient framework for understanding specialized functions within LLMs without the need for extensive model adjustments.

\section{Method}

\begin{table*}[ht!]
\caption{Summary of models, including their LLM family, layer-head architecture, resolution, training strategy, and use of visual tokens.} 
\label{table:model}
\centering
\scalebox{0.95}{
\begin{tabular}{ l|l|l|l|l|l }
\toprule
\textbf{Model} & \textbf{LLM Family} & \textbf{Layer-Head} & \textbf{Resolution} & \textbf{Training Strategy} & \textbf{Visual Tokens} \\ 
\midrule
vip-phi-3-3.8B & Phi-3 & $24 \times 32$ & $336 \times 336$ & frozen vision encoder & 576 \\
\midrule
1.6-mistral-7B & Mistral-v0.2 & $32 \times 32$ & Dynamic Res & full model trainable &$576 \times 1 \sim 4$ \\
\midrule
vip-llama-3-8B & Llama-3 & $24 \times 32$ & $336 \times 336$ & frozen vision encoder & 576 \\ 
\midrule
1.5-7B & \multirow{6}{*}{Vicuna-v1.5} & $32 \times 32$ & $336 \times 336$ & frozen vision encoder & 576 \\ 
1.6-vicuna-7B & & $32 \times 32$ & Dynamic Res & full model trainable &$576 \times 1 \sim 4$ \\
vip-7B & & $32 \times 32$ & $336 \times 336$ & frozen vision encoder & 576 \\  
\cmidrule(lr){1-1} \cmidrule(lr){3-6}
1.5-13B & & $40 \times 40$ & $336 \times 336$ & frozen vision encoder & 576 \\
1.6-vicuna-13B & & $40 \times 40$ & Dynamic Res & full model trainable &$576 \times 1 \sim 4$ \\
vip-13B & & $40 \times 40$ & $336 \times 336$ & frozen vision encoder & 576 \\
\bottomrule
\end{tabular}}
\end{table*}

\subsection{Experiment Setup}
\begin{figure}
    \centering
    \includegraphics[width=0.9\linewidth]{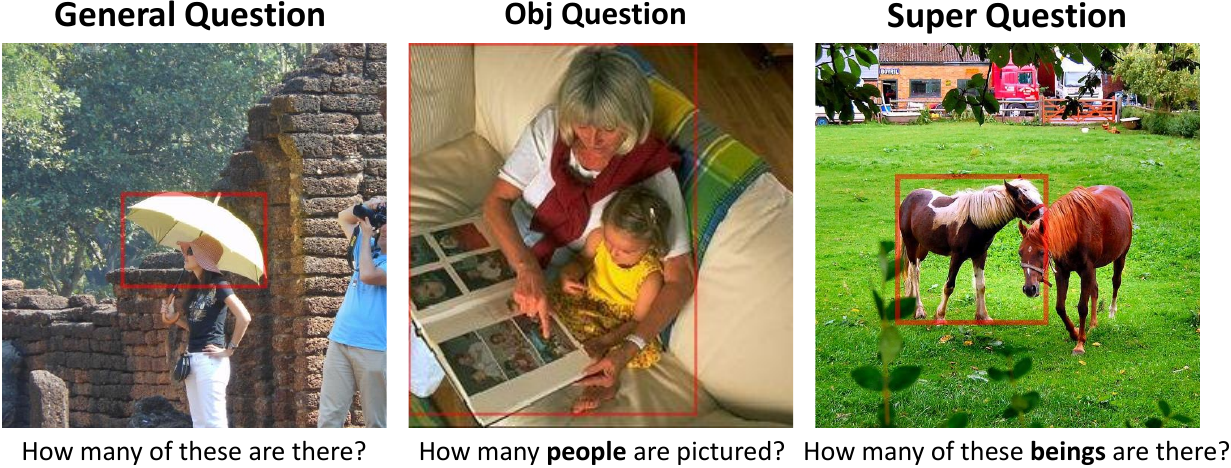}
    \caption{Example entries in the PointQA Dataset}
    \vspace{-5mm}
    \label{fig:point}
\end{figure}
We begin by selecting the specialized dataset, the PointQA dataset~\citep{mani2020point}, mainly for three reasons:
\begin{enumerate*}[label=(\alph*)]
\item Diverse Question Types: PointQA includes three categories of questions, each designed to vary in complexity and focus (as shown in Fig.\ref{fig:point}): \textit{General Questions:} These prompt the model to identify objects within specified regions and count instances (e.g., “How many objects are there?”), testing the model’s ability to recognize and count objects. \textit{Object-Specific Questions:} These directly identify the objects of interest, such as “How many people are in the picture?” This type assesses the model’s ability to detect and count specific objects within a scene. \textit{Super Questions:} These questions lack explicit object names (e.g., “How many beings are there?”), requiring the model to interpret the scene broadly to infer the correct subject.
\item Region-Specific Questioning and Visual Prompts: PointQA designates specific image regions relevant to each question, which serves two purposes. First, it allows us to quantitatively assess whether the model is focusing on the correct areas, ensuring accurate response generation. Second, by highlighting relevant areas (e.g., with a circle or box), we create visual prompts that guide the model’s attention to specific regions within the image.
\item Controlled Response: By limiting answers to a single numerical token (1-7), PointQA reduces the risk of hallucinated responses, ensuring that answers are directly tied to the visual input rather than influenced by model biases or prior content generation.
\item Look Twice: Consider questions that require spatial reasoning. A natural example is shown on the left in Fig\ref{fig:point}. The model needs to identify the relevant object in the image first and then use this information to attend to the entire image effectively.
\end{enumerate*}
\subsection{Model Selection and Configuration}

We examine how different LLM sizes, families, and visual-encoder combinations affect visual perception capabilities, focusing on integration methods ranging from simpler MLP adapters to more complex multi-layer CLIP representations, as seen in VIP-LLaVA.
Specifically, we evaluate nine variants of MLLMs, as shown in Table\ref{table:model}, with visual adapters across different configurations: LLaVA 1.5, which uses MLP layers as adapters to process visual input; VIP-LLaVA, based on LLaVA 1.5 but incorporating multiple intermediate representations from CLIP to enhance visual processing; and LLaVA 1.6, which increases the image token capacity from 576 to approximately 2304, improving benchmark performance at the cost of increased token numbers ($\sim$1.5k tokens/image), leading to higher computational costs. Note that we save \textit{llava} in all model names to optimize space.

\subsection{Extract Attention Weight}

In the PointQA dataset, the target token for each output is a single digit (ranging from 1 to 7). We analyze the attention weights, denoted as $\alpha_{l,h,j}$, which represent the probability that the target token attends to each input token $x_j$ for a specific attention head $h$ in layer $l$ of the 7b model. Here, both $l$ and $h$ range from 0 to 32.
The attention weights for each head form a $1 \times n$ vector, where $n$ is the length of the input sequence plus one (to account for the inclusion of the $\langle s \rangle$ start-of-sequence token). This structure minimizes interference from previously decoded tokens, meaning the target token’s information is derived directly from the input sequence.
Unlike previous studies that focus on attention flow~\cite{Abnar_2020} or relevance scores~\cite{Chefer_2021} to trace connections between input and output tokens, our work emphasizes how individual attention heads attend specifically to image tokens during decoding. This approach provides insights into how attention heads focus on the input image tokens.

\section{Basic Properties of Visual Heads}
\begin{figure*}[ht]
  \centering
  \includegraphics[width=\textwidth]{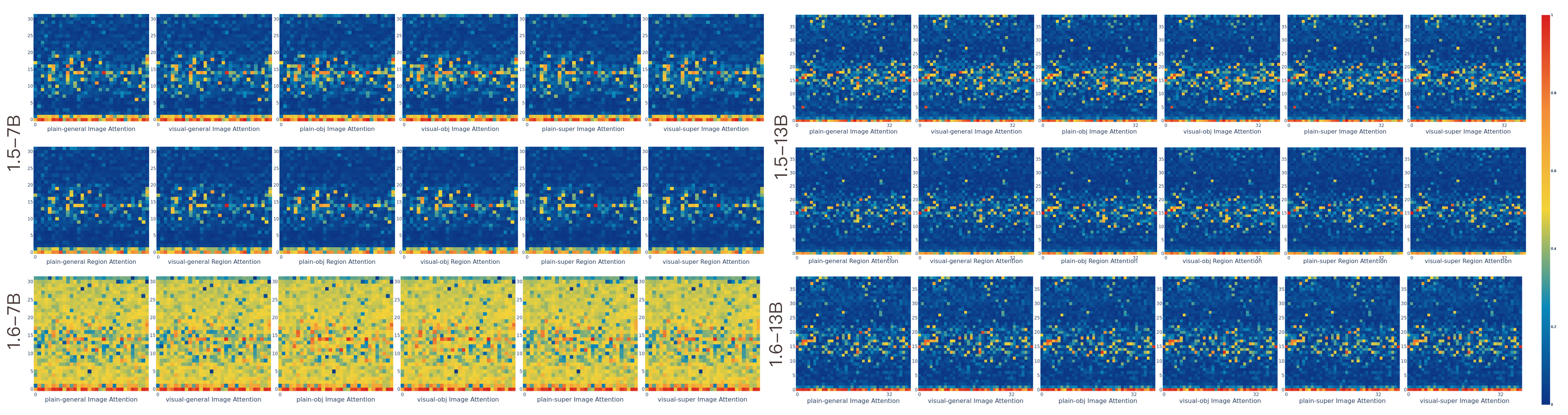}
  \caption{The image heatmap visualizes the total attention weight across image tokens and image region tokens. Attention is concentrated in specific layers, particularly in the early and middle layers. Comparing visual and plain generative tasks, we observe that the bounding box does not alter the attention head patterns. However, when comparing with a plain object prompt, including the object name in the question prompt activates additional attention heads not triggered by the visual prompt, suggesting that the attention heads exhibit dynamic activation based on the context—whether visual or linguistic. This highlights their ability to adjust their function and behavior in response to changing inputs. Further comparison between versions 1.6 and 1.5 demonstrates an improvement in image attention across all layers in version 1.6. However, this pattern is not as evident in the 1.6 13B model. The region token attention is omitted in 1.6 due to the more complex handling of the input image, making it challenging to track bbox token indices. Additionally, we see that the visual prompt does not improve the attention head’s focus on specific regions, as evidenced by comparing the first and second rows of the heatmap.}
  \label{fig:heatmap}
  \vspace{-2mm}
\end{figure*}

\begin{figure}[ht]
  \centering
  \includegraphics[width=0.375\textwidth]{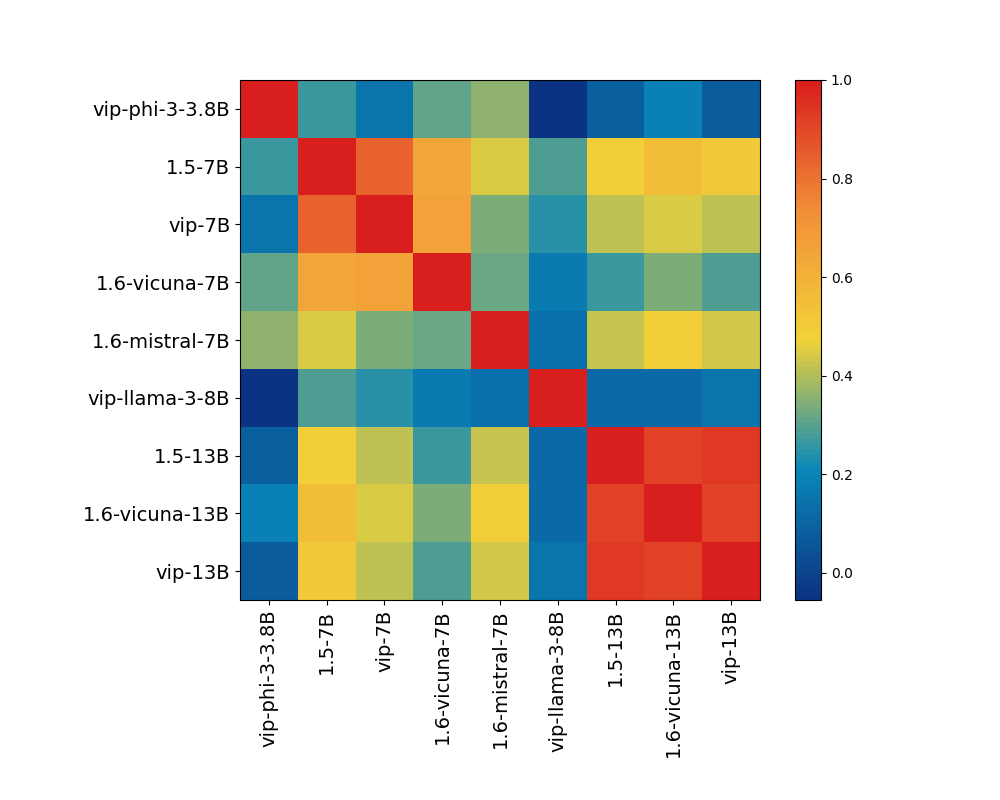}
  \vspace{-3mm}
  
  \caption{The visual heads of models within the same family exhibit a strong correlation, meaning that models of the same type typically share the same set of visual heads. In contrast, the visual heads of models from different families are distinctly different.}
  \label{fig:sim}
  \vspace{-3mm}
\end{figure}

\subsection{Attention Weight Analyze}
We conducted 6 controlled experiments across various data and question configurations, varying two key aspects: (1) attention guidance using visual cues, such as color bounding boxes (plain vs. visual), and (2) question types (general, object, and super-object queries). In each experiment, we analyzed the attention weights $\alpha_{l,h,j}$ for each model, with a particular focus on patterns in attention allocation.

For each sample, we recorded attention weights across layers $l$, heads $h$, yielding an array of dimensions $l \times h \times \text{seq\_len}$, where $\text{seq\_len}$ has an average length of 650 tokens in the PointQA dataset, including 576 tokens representing image regions. This allows us to partition attention analysis into two primary regions: the image token region and the question region token, which refers to the image token inside the bounding box from the question.  For each sample, we computed two key metrics:
\begin{enumerate*}[i]
\item The total attention weight in the image region.
\item The total attention weight in the question bounding box (bbox).
\end{enumerate*}
We then averaged these values across all 5,700 samples to obtain a comprehensive view of attention patterns across different model configurations, as visualized in Figure \(\ref{fig:heatmap}\). 
We observe several distinct patterns from the image.

\noindent \textbf{Attention Layer Localization}: Unlike the sparse and universal patterns found in retrieval heads, the visual heads focusing on the visual token displayed a unique pattern across layers. In the initial layers, attention weights for image tokens are high, indicating that early layers strongly focus on processing image information. Moving toward the middle layers, attention to visual tokens diminishes, showing a more diffused pattern. However, in the intermediate layers, attention to the image region increases again, before tapering off towards the final layers. Notably, a few heads regain significant attention at the output layer, suggesting a resurgence of focus on image regions at the final stages.

\noindent \textbf{Consistency Across Variants}: The above pattern was consistently observed in both the 7B and 13B models as well as across versions 1.5 and 1.6. Notably, the 1.6-7B variant showed the most pronounced attention on image tokens across all heads, suggesting enhanced visual focus in this setting.
When comparing the first and second columns, where the difference lies in whether a bounding box is drawn on the image, we observe that the relative changes are minimal. This suggests that adding a visual prompt does not lead non-visual heads to focus more on the image. However, we do observe a relatively higher attention weight in general visual cases. Additionally, comparing the first, second, and third rows reveals that object-related questions do indeed activate certain heads that remain inactive in visual-general scenarios, allocating more attention to the image (such as head (30,12)). In contrast, super questions appear to have little effect on attention heads.

One noticeable change occurs with the 7B model from version 1.5 to 1.6, where almost all heads allocate more attention to the image. This change can be attributed, in part, to the extension of image tokens to 1.5k. For instance, in the Point dataset, the combined system and user prompt accounts for approximately 70 tokens, which is only about 4\% of the image token count.

\subsection{Are Attention Weight Consistent?}
We extracted attention weights to four additional popular MLLM benchmarks: MMBench~\cite{liu2025mmbench}, POPE~\cite{li2023evaluating}, SEED~\cite{li2023seed}, and the recently introduced QBench~\cite{wu2023q}.

\begin{figure}[ht]
  \centering
  \includegraphics[width=0.45\textwidth]{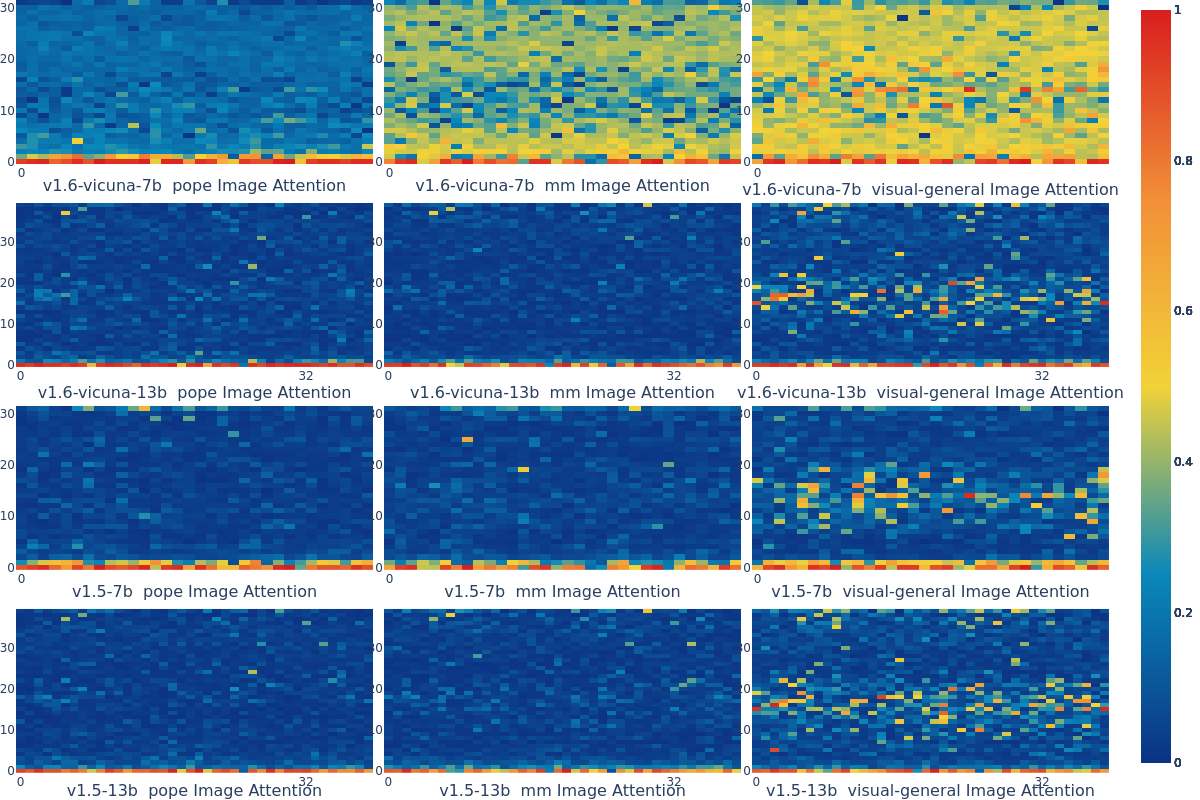}
  \caption{The attention weights demonstrate inconsistencies when applied to different datasets.}
  \label{fig:fail}
\vspace{-5mm}
\end{figure}

As shown in Fig.~\ref{fig:fail}, we observe that while different models exhibit similar attention patterns on Visual General subset, this consistency does not hold for datasets like MMBench and POPE. For these benchmarks, distinctive patterns emerge primarily in the first few layers. 
To investigate further, we conducted a thorough analysis of the hidden state outputs using Logit Lens~\cite{belrose2023eliciting}.
This analysis revealed that in datasets like POPE and MMBench, attention and logit progression stabilize after the initial layers. In contrast, for Visual General, attention weights and logits continue to evolve through the middle layers. This suggests that the answers require more refinement, as the dataset demands “look twice” reasoning, whereas the questions in POPE and MMBench are more straightforward.

Additionally, we focused on the behavior of attention heads in these benchmarks. 
A novel property emerged in the visual heads during our investigation of the hidden states at each layer: in the early layers, attention heads tend to focus on the entire image token, while in the later layers, attention increasingly concentrates on specific regions of interest within the image. This shift in focus may help explain the distinct behavior observed across different datasets.

\subsection{Concentration}
Based on our observations, we propose an enhance metric to more effectively capture attention head behavior across different datasets. Specifically, we recommend not only using attention weights but also incorporating a concentration score as a complementary dimension. This concentration score quantifies how narrowly or broadly a model head focuses on particular regions within an image as it processes each layer. Together, these metrics form a two-dimensional representation that offers a more comprehensive view of the model’s attention patterns. 
By using this attention-weight and concentration-score matrix, we can more accurately quantify and compare the model head behavior across diverse datasets, allowing us to identify dataset-specific characteristics and adaptations in attention dynamics.
The score is based on the entropy of the attention distribution across tokens. 
Given an attention vector \(\boldsymbol{\alpha}_{l,h,j}\), the entropy \(\mathcal{H}\) is calculated as:

$$
\mathcal{H} = -\sum_j \alpha_{l,h,j} \log(\alpha_{l,h,j} + \epsilon)
$$

where \(\epsilon\) is a small constant for numerical stability. To compute the concentration score \(\mathcal{C}\), we normalize this entropy by the maximum possible entropy \(\log_2(N + \epsilon)\), where \(N\) is the number of tokens:

$$
\mathcal{C} = 1 - \frac{\mathcal{H}}{\log_2(N + \epsilon)}
$$

The concentration score \(\mathcal{C}\) ranges from 0 to 1, with higher values indicating that the model's attention is more narrowly focused on specific parts of the input. We compute the \(\mathcal{C}\) and attention weight of each head and average these values over the entire dataset. We then generate a scatter plot to visualize the results, as shown in Fig\ref{fig:sumvcon}
\begin{figure*}[ht]
  \centering
  \includegraphics[width=0.9\textwidth]{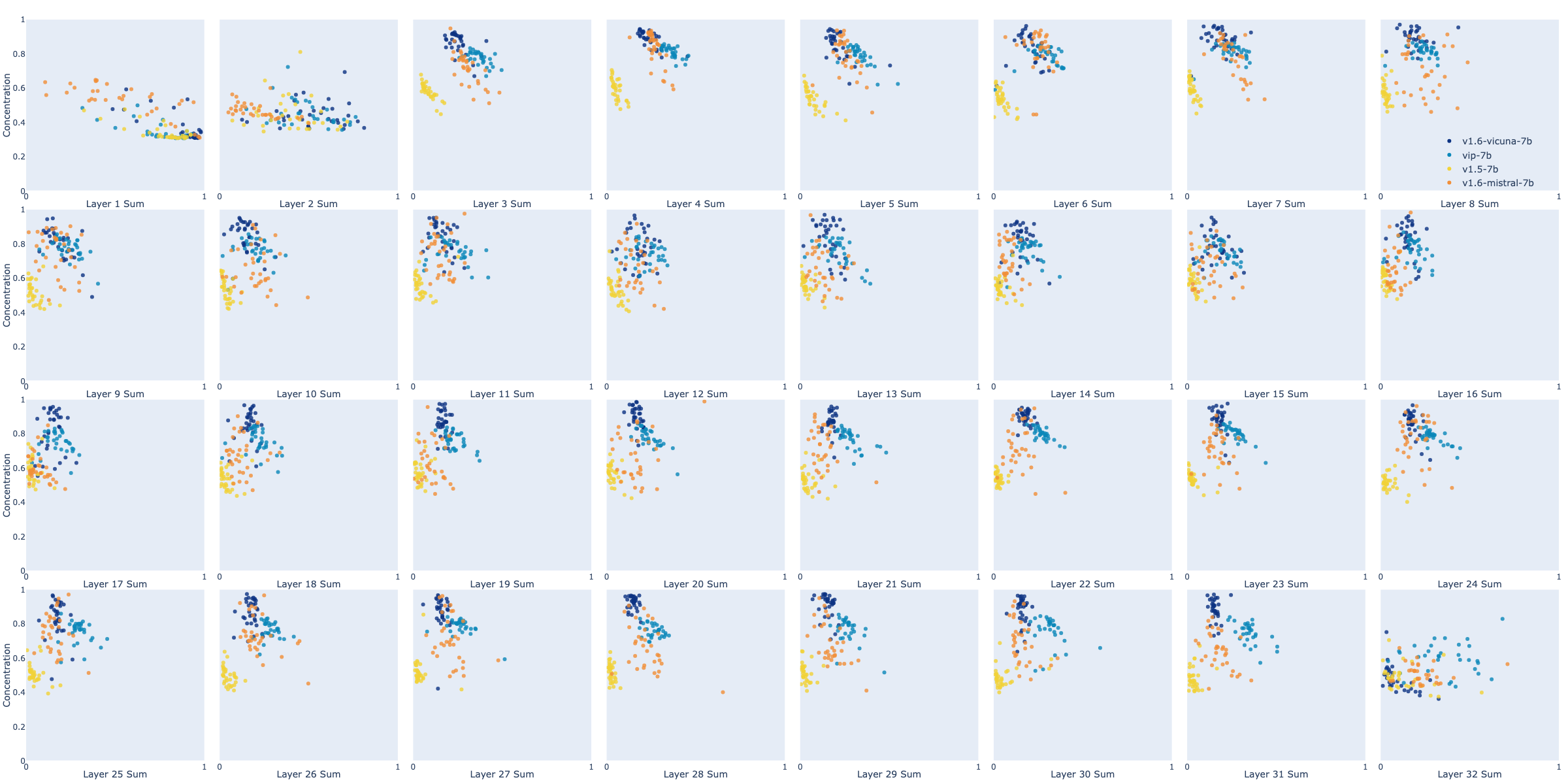}
  \caption{Visualization of Attention Weights and Concentration Scores across Different Models and Layers. Each subplot represents a distinct layer, with colors indicating various models, showcasing how attention allocation and concentration vary across layers.}
  \label{fig:7b}
\vspace{-1mm}
\end{figure*}

\subsection{Analysis of Weight vs. Concentration}
\begin{figure}
    \centering
    \includegraphics[width=1\linewidth]{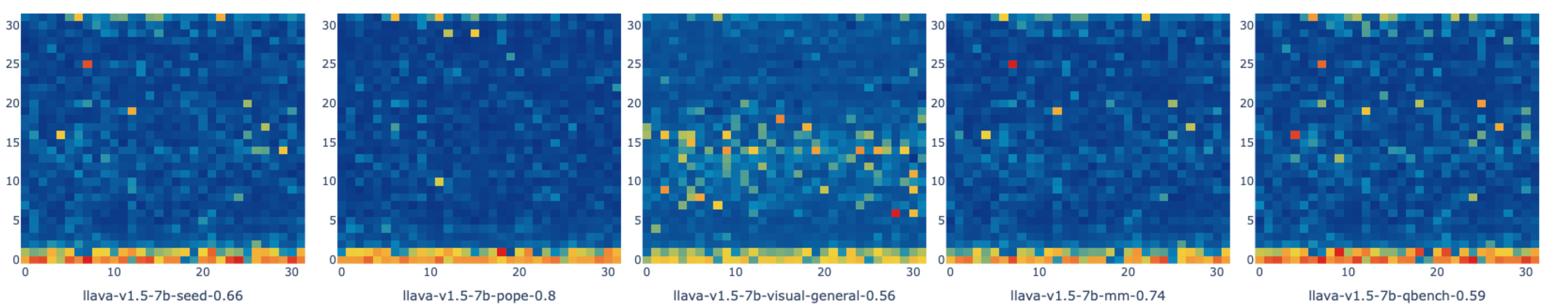}
    \caption{After applying the concentration score, we observe a similar pattern as seen when using only the attention weight. This consistency across all five datasets suggests that concentration and attention weight scores can effectively be used together to analyze attention behavior across datasets.}
    \label{fig:fix}
    \vspace{-1mm}
\end{figure}

This analysis essentially demonstrates how each attention head localizes its focus across the entire dataset. As shown in the figure, when comparing the plots across rows, we observe that heads with higher concentration or those focusing more on specific parts of the image tend to yield better performance compared to models that do not exhibit this behavior. Furthermore, we found that the 13B family of models displays very similar behavior across the five benchmarks.
As shown in Fig.~\ref{fig:sumvcon}, we observe a consistent pattern of model behavior within the dataset.
Moreover, we found that integrating concentration with attention weight still holds in our analysis of the distribution of heads across layers as show in Fig\ref{fig:fix}. We can still observe the distribution layer-wise. Additional figures detailing different models and datasets will be included in the supplementary.

\noindent \textbf{Correlation Analysis within Model Families} We evaluate the statistical correlations between different models by calculating the Pearson correlation coefficient, which is visualized using a combination of  Concentration and attention weight in Visual General subset. The findings indicate a strong correlation, with a Pearson correlation coefficient greater than 0.8, between the retrieval score distributions of base models and their respective variants. On the other hand, models from different families show a correlation coefficient below 0.1, highlighting the differences in their pretraining approaches.

\subsection{Head Detection Score}

In multimodal models, decoding language tokens does not yield a straightforward one-to-one correspondence with input tokens, necessitating an aggregated approach across token regions. From our observations, we visualized attention dynamics within the 7b model (see Fig. \ref{fig:7b}), uncovering intriguing patterns in visual token behavior across layers. In the first layer, attention is broadly distributed across the image, reflecting a wide, exploratory scan. By the second layer, attention begins to concentrate and move leftward, suggesting a refining process towards specific regions. This transition stabilizes in the middle layers, with a trend towards higher concentration and more precise allocation of attention. Comparing versions v1.5 and v1.6 of the 7b model, the advanced version (v1.6) maintains higher attention values and demonstrates greater focus, indicating improvements in visual content engagement. Our analysis highlights three key elements in attention dynamics: (1) attention weight, (2) concentration, and (3) the specific layer. Early stages show uniform attention distribution, while later layers exhibit increasingly focused attention, suggesting refinement as the model progresses.To detect heads accordingly, our score function is defined over the layer index, attention weight, and attention concentration:
$$
\text{Score}(l, h, j) = \sum_{\text{part of } j \in I} \boldsymbol{\alpha}_{l, h, j} \cdot \left(1 + \mathcal{C} \cdot \text{func}(l)\right)
$$
where $\text{func}(l) = \frac{k}{l + \epsilon} + a e^{-b l}$ and \( \boldsymbol{\alpha}_{l, h, j} \) represents the attention weights of the head indexed by \( h \) at layer \( l \) for a specific part of the attention context \( j \). The function \( \text{func}(l) \) modulates the layer contribution, ensuring: (1) Higher scores for lower layers (\( l \approx 0 \)), offsetting the smaller \( \mathcal{C} \). (2) Gradual reduction of influence as \( l \) increases, reflecting decreasing relevance in deeper layers.

The term \( \frac{k}{l + \epsilon} \) amplifies contributions from lower layers, with \( \epsilon \) ensuring stability. Together, these terms adaptively highlight attention heads with meaningful contributions while maintaining a layer-dependent balance.

\begin{table}[h!]
\centering
\resizebox{0.48\textwidth}{!}{%
\begin{tabular}{|l|c|c|c|c|c|c|}

\hline
\multirow{2}{*}{\textbf{Category}} & \multicolumn{3}{c|}{\textbf{\textbf{1.6-vicuna-7b}}}& \multicolumn{3}{c|}{\textbf{\textbf{1.5-7b}}} \\
\cline{2-7}
 & \textbf{MMBench} & \textbf{POPE} & \textbf{QBENCH} & \textbf{MMBench} & \textbf{POPE} & \textbf{QBENCH}\\
\hline
No mask acc & 75.44 & 85.27 & 63.73& 77.64	&85.54	&66.64 \\
\hline
early top 10 & \textcolor[rgb]{0.043,0.362,0.618}{9.72 $\downarrow$} & \textcolor[rgb]{0.040,0.483,0.697}{16.07 $\downarrow$} & \textcolor[rgb]{0.047,0.213,0.522}{1.23 $\downarrow$} & \textcolor[rgb]{0.044,0.348,0.610}{8.34 $\downarrow$} & \textcolor[rgb]{0.043,0.389,0.636}{10.39 $\downarrow$} & \textcolor[rgb]{0.043,0.362,0.618}{9.18 $\downarrow$} \\
\hline
early top 20 & \textcolor[rgb]{0.453,0.667,0.498}{27.64 $\downarrow$} & \textcolor[rgb]{0.949,0.768,0.220}{41.47 $\downarrow$} & \textcolor[rgb]{0.159,0.572,0.662}{21.22 $\downarrow$} & \textcolor[rgb]{0.490,0.679,0.477}{27.94 $\downarrow$} & \textcolor[rgb]{0.600,0.715,0.415}{29.71 $\downarrow$} & \textcolor[rgb]{0.122,0.560,0.683}{20.72 $\downarrow$} \\
\hline
early top 30 & \textcolor[rgb]{0.918,0.422,0.188}{61.65 $\downarrow$} & \textcolor[rgb]{0.899,0.333,0.167}{66.1 $\downarrow$} & \textcolor[rgb]{0.949,0.671,0.220}{48.56 $\downarrow$} & \textcolor[rgb]{0.910,0.386,0.179}{62.34 $\downarrow$} & \textcolor[rgb]{0.946,0.547,0.217}{55.8 $\downarrow$} & \textcolor[rgb]{0.949,0.682,0.220}{46.7 $\downarrow$} \\
\hline
early top 50 & \textcolor[rgb]{0.871,0.207,0.138}{70.74 $\downarrow$} & \textcolor[rgb]{0.851,0.118,0.118}{75.17 $\downarrow$} & \textcolor[rgb]{0.926,0.458,0.196}{60.47 $\downarrow$} & \textcolor[rgb]{0.863,0.171,0.130}{71.61 $\downarrow$} & \textcolor[rgb]{0.871,0.207,0.138}{69.91 $\downarrow$} & \textcolor[rgb]{0.949,0.574,0.220}{54.56 $\downarrow$} \\
\hline
early others 20 & \textcolor[rgb]{0.047,0.213,0.522}{1.21 $\downarrow$} & \textcolor[rgb]{0.046,0.240,0.540}{2.82 $\downarrow$} & \textcolor[rgb]{0.047,0.200,0.514}{0.74 $\downarrow$} & \textcolor[rgb]{0.046,0.240,0.540}{2.38 $\downarrow$} & \textcolor[rgb]{0.046,0.227,0.531}{2.13 $\downarrow$} & \textcolor[rgb]{0.047,0.200,0.514}{0.1 $\downarrow$} \\
\hline
early others 50 & \textcolor[rgb]{0.043,0.375,0.627}{10.17 $\downarrow$} & \textcolor[rgb]{0.040,0.483,0.697}{16.07 $\downarrow$} & \textcolor[rgb]{0.047,0.213,0.522}{1.23 $\downarrow$} & \textcolor[rgb]{0.045,0.294,0.575}{5.88 $\downarrow$} & \textcolor[rgb]{0.046,0.240,0.540}{2.8 $\downarrow$} & \textcolor[rgb]{0.045,0.308,0.583}{6.28 $\downarrow$} \\
\hline
early others 80 & \textcolor[rgb]{0.490,0.679,0.477}{28.36 $\downarrow$} & \textcolor[rgb]{0.949,0.768,0.220}{41.47 $\downarrow$} & \textcolor[rgb]{0.159,0.572,0.662}{21.22 $\downarrow$} & \textcolor[rgb]{0.949,0.768,0.220}{41.29 $\downarrow$} & \textcolor[rgb]{0.949,0.563,0.220}{55.41 $\downarrow$} & \textcolor[rgb]{0.048,0.536,0.724}{18.89 $\downarrow$} \\
\hline
early others 100 & \textcolor[rgb]{0.922,0.440,0.192}{61.47 $\downarrow$} & \textcolor[rgb]{0.899,0.333,0.167}{65.56 $\downarrow$} & \textcolor[rgb]{0.949,0.671,0.220}{48.56 $\downarrow$} & \textcolor[rgb]{0.902,0.350,0.171}{63.96 $\downarrow$} & \textcolor[rgb]{0.851,0.118,0.118}{74.08 $\downarrow$} & \textcolor[rgb]{0.938,0.512,0.208}{57.09 $\downarrow$} \\
\hline

\end{tabular}
}
\caption{Performance drops for both models, measured in MMBench, POPE, and QBENCH. The results highlight the performance impact under different early stage masking conditions.}
\end{table}


\begin{figure}
    \centering
    \includegraphics[width=1\linewidth]{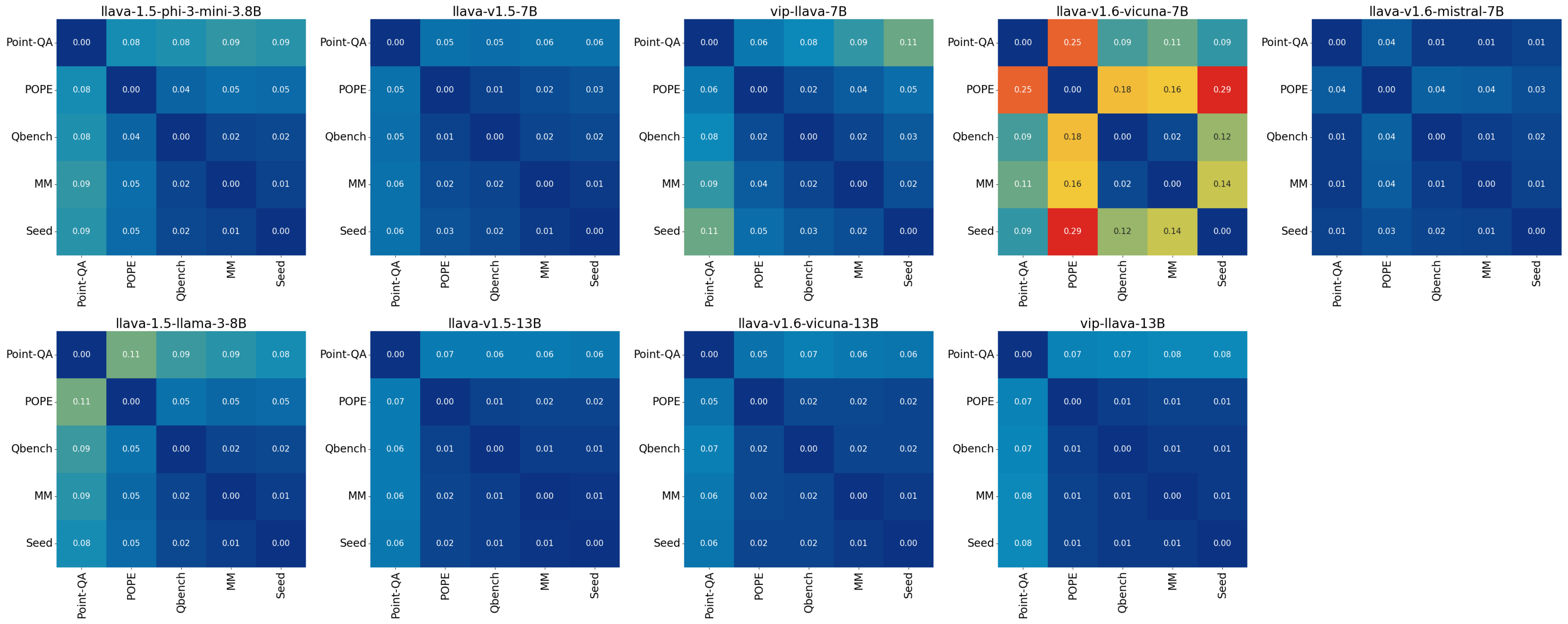}
    \caption{EMD (Earth Mover’s Distance) score matrices over 5 benchmark datasets for each model.}
    \label{fig:EMD}
\vspace{-1mm}
\end{figure}

\section{Application}

\subsection{Impact of heads}

In our study, we divided each model into three stages, each representing a third of the model’s layers. We observed that heads tend to cluster in certain layers; therefore, we divided the layers accordingly to analyze the impact systematically. We conducted two experiments: \textit{Top Head Selection}: We sampled the top 20 heads from each stage, aiming to assess the impact of these heads on performance metrics. Our hypothesis was that these top heads play a critical role in the model’s visual understanding abilities.
\textit{Head Removal Impact}: After observing that removing heads in the early stages led to the most substantial performance drop, we further analyzed the effect of incrementally removing heads to determine how many heads can be removed before significant performance degradation occurs.

The results show a  clear performance drop when heads from the early layers are removed. Removing the top 20 heads from early layers reduces performance by roughly 20–40\%, emphasizing the role of early-stage heads in fundamental feature extraction.
Meanwhile, the performance drop from mid and late stages is less pronounced, with the late stages showing only minor performance decreases.
Incrementally increasing the number of removed heads in early stages shows a near-linear decrease in model performance, with accuracy dropping by as much as 50\% when 50 heads are removed. This indicates that while some redundancy exists, a threshold exists after which too many removed heads critically weaken model understanding.
Removing up to 20 heads in the early stage still allows the model to retain significant accuracy. However, exceeding this number (e.g., 30 or 50 heads) begins to result in sharp declines, showing that these heads collectively perform essential functions that cannot be compromised without noticeable performance loss.
We also conducted experiments focusing on the removal of mid- and late-stage heads.
In these experiments, we found that up to 100 heads can be removed from the late stage without significantly impacting performance. This suggests that after the logical structures stabilize in the mid layers, the results remain largely unchanged in the last few layers. This finding is an important indication that unnecessary computations in the late layers can be pruned during inference. As token lengths increase, efficient inference becomes increasingly critical, and this pruning approach offers a promising direction to optimize computational efficiency.

\begin{table}[h!]
\centering
\resizebox{0.48\textwidth}{!}{%
\begin{tabular}{|l|c|c|c|c|c|c|}

\hline
\multirow{2}{*}{\textbf{Category}} & \multicolumn{3}{c|}{\textbf{\textbf{1.6-vicuna-7b}}}& \multicolumn{3}{c|}{\textbf{\textbf{1.5-7b}}} \\
\cline{2-7}
 & \textbf{MMBench} & \textbf{POPE} & \textbf{QBENCH} & \textbf{MMBench} & \textbf{POPE} & \textbf{QBENCH}\\
\hline
No mask acc & 73.61 & 79.67 & 59.09 & 75.44 & 85.27 & 63.37\\
\hline
early top 20 & \textcolor[rgb]{0.871,0.207,0.138}{27.93 $\downarrow$} & \textcolor[rgb]{0.851,0.118,0.118}{29.71 $\downarrow$} & \textcolor[rgb]{0.949,0.650,0.220}{19.72 $\downarrow$} & \textcolor[rgb]{0.902,0.350,0.171}{24.43 $\downarrow$} & \textcolor[rgb]{0.851,0.118,0.118}{28.21 $\downarrow$} & \textcolor[rgb]{0.949,0.574,0.220}{20.86 $\downarrow$} \\
\hline
mid top 20 & \textcolor[rgb]{0.490,0.679,0.477}{11.01 $\downarrow$} & \textcolor[rgb]{0.416,0.655,0.518}{10.58 $\downarrow$} & \textcolor[rgb]{0.526,0.691,0.456}{11.58 $\downarrow$} & \textcolor[rgb]{0.673,0.738,0.374}{11.86 $\downarrow$} & \textcolor[rgb]{0.949,0.779,0.220}{15.34 $\downarrow$} & \textcolor[rgb]{0.453,0.667,0.498}{10.39 $\downarrow$} \\
\hline
late top 20 & \textcolor[rgb]{0.043,0.389,0.636}{4.36 $\downarrow$} & \textcolor[rgb]{0.046,0.254,0.549}{1.44 $\downarrow$} & \textcolor[rgb]{0.047,0.200,0.514}{0.29 $\downarrow$} & \textcolor[rgb]{0.040,0.510,0.714}{6.56 $\downarrow$} & \textcolor[rgb]{0.048,0.536,0.724}{7.21 $\downarrow$} & \textcolor[rgb]{0.041,0.442,0.671}{5.14 $\downarrow$} \\
\hline
early random 20 & \textcolor[rgb]{0.044,0.335,0.601}{3.24 $\downarrow$} & \textcolor[rgb]{0.045,0.294,0.575}{2.3 $\downarrow$} & \textcolor[rgb]{0.045,0.267,0.557}{1.78 $\downarrow$} & \textcolor[rgb]{0.044,0.348,0.610}{3.3 $\downarrow$} & \textcolor[rgb]{0.043,0.389,0.636}{3.95 $\downarrow$} & \textcolor[rgb]{0.044,0.348,0.610}{3.25 $\downarrow$} \\
\hline
mid random 20 & \textcolor[rgb]{0.047,0.213,0.522}{0.51 $\downarrow$} & \textcolor[rgb]{0.045,0.267,0.557}{1.55 $\downarrow$} & \textcolor[rgb]{0.046,0.240,0.540}{1.18 $\downarrow$} & \textcolor[rgb]{0.045,0.294,0.575}{1.98 $\downarrow$} & \textcolor[rgb]{0.044,0.335,0.601}{2.91 $\downarrow$} & \textcolor[rgb]{0.045,0.294,0.575}{2.15 $\downarrow$} \\
\hline
late random 20 & \textcolor[rgb]{0.046,0.227,0.531}{0.62 $\downarrow$} & \textcolor[rgb]{0.046,0.254,0.549}{1.22 $\downarrow$} & \textcolor[rgb]{0.047,0.200,0.514}{0.16 $\downarrow$} & \textcolor[rgb]{0.046,0.240,0.540}{0.95 $\downarrow$} & \textcolor[rgb]{0.046,0.240,0.540}{0.96 $\downarrow$} & \textcolor[rgb]{0.045,0.267,0.557}{1.5 $\downarrow$} \\
\hline
\end{tabular}
}
\caption{Performance drops for both models, measured in MMBench, POPE, and QBENCH. The results highlight the performance impact under all stages masking conditions.}
\vspace{-2mm}
\end{table}

\subsection{Attention Head for Model Behavior Analysis}
Using EMD (earth mover's distance), we calculated the correlation between five benchmarks for the same models, as shown in Fig.~\ref{fig:EMD}. In this figure, \textbf{llava-v1.6-vicuna-7B} show notably low correlations between their performance on the POPE benchmark and other benchmarks. Our findings suggest that examining a model’s attention and concentration patterns could help predict its performance on related benchmarks. 


\section{Conclusion}

Our work reveals that visual heads in LLMs are concentrated in specific layers, primarily early and middle stages, playing a key role in processing visual tokens. By conducting comprehensive experiments, we found that visual heads demonstrate dynamic activation, high concentration correlates with better performance, and early-stage heads are critical for visual understanding. Late-stage heads contribute minimally, offering potential for pruning to improve computational efficiency. Larger models show enhanced consistency and adaptability across datasets.


\section{Acknowledgment}
This work was supported in part by the Defense Advance Research Projects Agency under contract number HR00112220003. The content is solely the responsibility of the authors and does not necessarily represent the official views of the funding agencies; no official endorsement should be inferred.
{
\cleardoublepage
    \small
    \bibliographystyle{ieeenat_fullname}
    \bibliography{main}
}

\end{document}